\documentclass[11pt]{article}

\usepackage[preprint]{acl}
\usepackage{microtype}
\usepackage{graphicx}
\usepackage{subcaption}
\usepackage{booktabs}
\usepackage{multirow}
\usepackage[table]{xcolor}
\usepackage{adjustbox}
\usepackage{float}
\usepackage{placeins}
\usepackage[most]{tcolorbox}
\usepackage{enumitem}
\usepackage{tcolorbox}
\usepackage{amssymb}
\tcbuselibrary{breakable}
\usepackage{pifont}

\usepackage{times}
\usepackage{latexsym}

\usepackage[T1]{fontenc}

\usepackage[utf8]{inputenc}

\usepackage{microtype}

\usepackage{inconsolata}

\usepackage{graphicx}

\title{Beyond Trajectory Rewards: Step-level Credit Assignment for Agentic Search via Graph Modeling}

\author{
Yuchen Liu\textsuperscript{1}\Thanks{Equal contribution.}
\quad
Yingjie Feng\textsuperscript{2}\footnotemark[1]
\quad
Lixiong Qin\textsuperscript{1}\footnotemark[1]
\\
Jiasi Chen\textsuperscript{1}
\quad
Jianing Yu\textsuperscript{1}
\quad
Sheng Gao\textsuperscript{1}
\quad
Sheng Yang\textsuperscript{2}\Thanks{Corresponding author.}
\quad
Weiran Xu\textsuperscript{1}\footnotemark[2]
\\
\textsuperscript{1}Beijing University of Posts and Telecommunications
\\
\textsuperscript{2}Li Auto Inc.
\\
\texttt{ycliu@bupt.edu.cn}
\quad
\texttt{yangsheng3@lixiang.com}
\quad
\texttt{xuweiran@bupt.edu.cn}
}

\begin{document}
\maketitle
\begin{abstract}
In Agentic Search, trajectory-level outcome rewards fail to quantify the behavioral contributions of individual steps, while existing step-level reward methods typically rely on costly tree sampling. We view world knowledge as a latent world graph and each IS task as search within a latent task graph, where effective steps should make graph progress toward the answer node. Based on this prior, we propose Graph-Distance Contribution Reward (GDCR), a step-level process reward that scores newly-retrieved and newly-cited entities by their distance to the answer node in a training-time Entity-Relation (ER) graph. We further propose Step Advantage Policy Optimization (SAPO), which converts GDCR into step-level advantages and combines them with trajectory-level outcome advantages. Experiments on four challenging benchmarks validate the effectiveness of our method.
\end{abstract}

\section{Introduction}

\begin{figure}[t]
\centering
\includegraphics[width=\columnwidth]{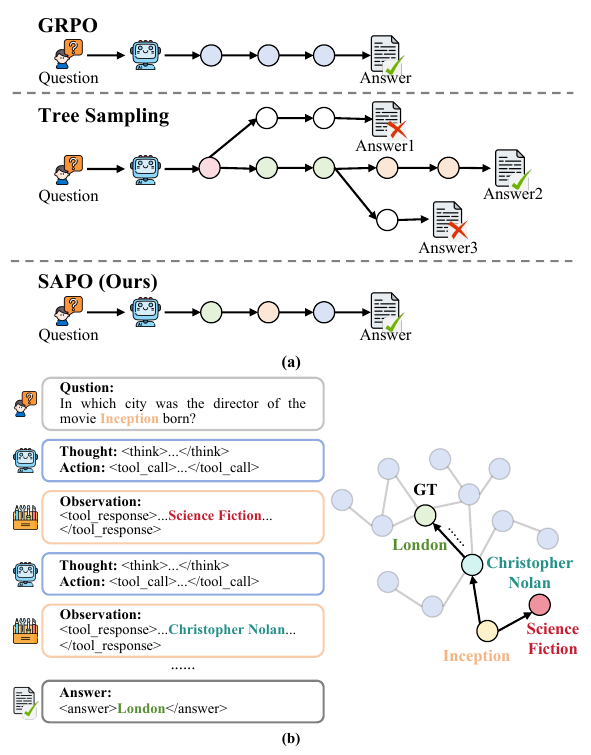}
\caption{
(a) Comparison between our proposed SAPO and other RL methods, where steps indicated with different colors have different advantage values. 
GRPO neglects the differences in the contribution of each step, whereas Tree Sampling-based step-level policy optimization incurs substantial computational overhead in estimating step contributions.
(b) Our method is based on the prior that the more a retrieved entity advances the search agent toward the answer node in the latent task graph, the greater its contribution to obtaining a correct solution.
}
\label{fig:single_col}
\vspace{-10pt}
\end{figure}

In recent years, LLM-based agents have made tremendous progress \cite{yao2022react,wei2022chain,shinn2023reflexion}.
Based on this, Agentic Search has garnered significant attention from both academia and industry, with LLM-based agents being applied to tackling complex Information Seeking (IS) tasks.
Unlike traditional retrieval methods, search agents iteratively interact with external knowledge bases and update their thoughts based on retrieved results \cite{KimiResearcher2025,OpenAIDeepResearch2025}. This enables them to solve knowledge-intensive problems beyond single-turn inference.
Recent studies further leverage RL \cite{li2025searcho1,jin2025search,zheng2025deepresearcher} to improve agents' search capabilities.

Currently, mainstream RL methods typically employ trajectory-level rewards, with Group Relative Policy Optimization (GRPO) and its variants occupying a dominant position \cite{shao2024deepseekmath,yu2025dapo,zheng2025group}. These methods determine rewards based on the correctness of the final answer and uniformly assign the signal across all steps in the trajectory.
In Agentic Search \cite{li2025searcho1,song2025r1,team2025tongyi}, although trajectory-level rewards have achieved significant success, they overlook the fact that different steps in a multi-turn search process contribute differently to the final success.
As a result, they fail to accurately measure the quality of individual steps, making it difficult to optimize critical retrieval and citation behaviors in a targeted manner.
Therefore, this paper aims to introduce step-level rewards into the policy optimization framework.

Existing step-level policy optimization methods typically rely on sampling-based approaches to assign rewards to each step \cite{dong2025agentic,dong2025agenticE,ji2025tree,yang2025treerpo,hou2025treerl}. They usually employ tree sampling to explore multiple continuations and back-propagate final rewards from leaf nodes to intermediate steps.
However, for complex IS tasks, agent trajectories often extend to dozens of steps, and a large branching factor is required to avoid missing critical decision points.
This leads to excessive computational overhead and severely limits training efficiency.
This naturally raises a key question: can we obtain effective step-level credit assignment without relying on expensive tree sampling?

To answer this question, we propose an intuitive answer-grounded prior:
world knowledge can be viewed as a large latent world graph, and each IS task corresponds to a latent task graph that connects query-conditioned entities to the answer node.
Agentic Search can then be viewed as gradually exploring this latent task graph, where an effective step should make graph progress toward the answer node.
As shown in Figure~\ref{fig:single_col}(b), for the question ``In which city was the director of the movie Inception born?'', retrieving ``Christopher Nolan'' bridges the query entity ``Inception'' and the answer ``London'', and thus deserves a high reward. In contrast, retrieving ``Science Fiction'', though semantically related to the movie, is logically distant from the answer and contributes little.
Based on this prior, we construct a training-time Entity-Relation (ER) graph for each Question-Answering (QA) pair as a proxy for the latent task graph, where nodes represent entities and edges represent semantic relationships.
We estimate the contribution of a graph-linked entity using its shortest-path distance to the answer node, and define one component of the step-level reward as the sum of the graph-distance contribution scores of newly-retrieved entities.

Furthermore, our analysis in Figure~\ref{fig:dist-percentage} reveals that some retrieved entities, despite being answer-relevant, may not be explicitly cited in later thoughts and thus fail to contribute to the final answer.
To address this issue, we further incorporate the graph-distance contribution scores of newly-cited entities in the current thought.
Similar to newly-retrieved entities, newly-cited entities closer to the answer node are more likely to move the search path toward the answer and therefore should receive higher reward scores.
Consequently, we propose the \textbf{Graph-Distance Contribution Reward (GDCR)}, which encourages agents to retrieve and cite entities closer to the answer node for more effective search.

Building on GDCR, we propose \textbf{Step Advantage Policy Optimization (SAPO)}. SAPO first normalizes and bounds GDCR signals within each trajectory, and then combines trajectory-level rewards with GDCR-based step-level rewards to formulate the advantage.
The trajectory-level reward preserves the final-answer objective, while the step-level reward reinforces critical behaviors and suppresses redundant ones.
As shown in Figure~\ref{fig:single_col}(a), SAPO alleviates reward sparsity in long-horizon IS tasks and avoids the prohibitive overhead of sampling-based step-level reward estimation.
Experiments on four challenging benchmarks, together with targeted analyses, validate the effectiveness and efficiency of SAPO.
Notably, SAPO outperforms ARPO, a representative tree-sampling step-level reward method, in both training efficiency and performance.

The main contributions of this paper can be summarized as follows:
\begin{itemize}
  \item We propose the answer-grounded graph-progress prior for Agentic Search, which views each IS task as exploration over a latent task graph within a latent world graph, and we use a training-time Entity-Relation (ER) graph as a proxy for the latent task graph to derive step-level credit from graph-distance progress toward the answer node.
  \item We propose \textbf{Graph-Distance Contribution Reward (GDCR)}, which assigns step-level credit to both newly-retrieved entities and newly-cited entities, encouraging the agent to retrieve and cite entities that make stronger graph progress toward the answer node.
  \item We propose \textbf{Step Advantage Policy Optimization (SAPO)}, which combines trajectory-level rewards with GDCR-based step-level rewards and achieves strong performance on multiple datasets.
\end{itemize}

\section{Preliminaries}
\label{sec:preliminaries}

We first formulate information-seeking tasks from the perspective of search agents, and then introduce the graph-augmented data setting that provides supervision for GDCR.

\subsection{Task Formulation}

\noindent\textbf{Information-Seeking Task.}
We consider complex information-seeking (IS) tasks where an agent is given a user query \(q\) and required to produce a final answer by collecting and reasoning over external information. Unlike single-hop retrieval tasks, complex IS tasks often require discovering multiple pieces of evidence and identifying their dependencies before arriving at the answer.

We view world knowledge as a latent world graph:
\begin{equation}
\mathcal{G}^{\mathrm{world}} = (\mathcal{V}^{\mathrm{world}}, \mathcal{E}^{\mathrm{world}}),
\end{equation}
where \(\mathcal{V}^{\mathrm{world}}\) denotes entities in world knowledge, and \(\mathcal{E}^{\mathrm{world}}\) denotes semantic or relational dependencies among them. For a query \(q\), the information needed to solve the task corresponds to a latent task graph:
\begin{equation}
\mathcal{G}_q = (\mathcal{V}_q, \mathcal{E}_q),
\end{equation}
where \(\mathcal{G}_q \subseteq \mathcal{G}^{\mathrm{world}}\), \(\mathcal{V}_q\) denotes the set of task-relevant entities, and \(\mathcal{E}_q\) denotes their semantic or relational dependencies. The answer corresponds to an answer node \(v_q^\ast \in \mathcal{V}_q\). Under this view, solving an IS task requires the agent to explore the latent task graph from query-conditioned entities toward the answer node. Therefore, an effective step should move the agent closer to \(v_q^\ast\), while redundant or irrelevant steps contribute little to solving the task.

\noindent\textbf{Search Agent.}
We formalize the search agent using the ReAct paradigm \cite{yao2022react}. Given a user query \(q\), an LLM-based agent, governed by a policy \(\pi_{\theta}\), incrementally collects evidence and produces a final response through multiple rounds of interaction with an external search environment.
Specifically, the agent executes a thought-action-observation loop in each iteration: it first generates a free-form thought \(\tau_t\) based on the current context, then selects an action \(a_t\), such as submitting a search query, and receives a corresponding observation \(o_t\) from the environment, typically consisting of retrieved document snippets. The interaction terminates when the agent selects \textless answer\textgreater{} as its action. A complete trajectory of \(T\) iterations is denoted as:
\begin{equation}
\mathcal{H}_{T} = \left( q, \tau_1, a_1, o_1, \ldots, \tau_{t}, a_{t}, o_{t}, \ldots, \tau_{T}, a_{T} \right).
\label{eq:Ht}
\end{equation}
where \(\tau_t\), \(a_t\), and \(o_t\) represent the thought, action, and observation at step \(t\). At each step, the thought and action are sampled from the policy \(\pi_\theta\) based on the historical context \(\mathcal{H}_{t-1}\). Additional format details can be found in Appendix~\ref{format}.

\noindent\textbf{Tool Design.}
Our agent is equipped with a single search tool: a search engine that supports multiple queries and returns the top-5 relevant results for each query, including the corresponding titles, abstracts, and URLs.

\subsection{Data Synthesis}

Large-scale and high-quality data are essential for training search agents with reinforcement learning \cite{wu2025webdancer,gao2508beyond}. In addition to QA pairs and search trajectories, our synthesis process constructs a training-time ER graph \(\hat{\mathcal{G}}_q\) for each query-answer pair, with task-relevant entities, relations, and answer-node annotations. This graph serves as an observable proxy for the latent task graph and makes it possible to estimate step-level contribution through graph distance. More discussion of graph construction and quality control can be found in Appendix~\ref{data}.

\section{Methodology}
\label{sec:methodology}

\begin{figure*}[t]
  \centering
  \includegraphics[width=\textwidth]{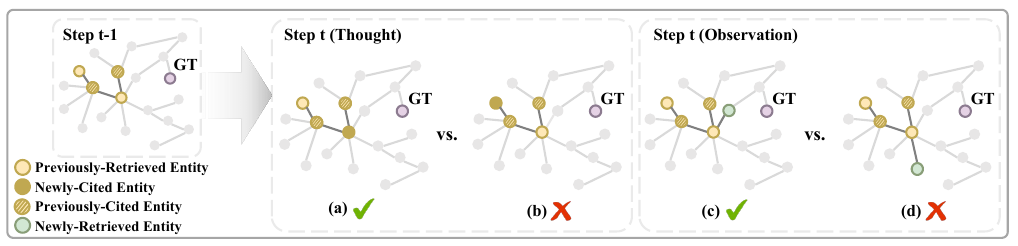}
  \caption{Illustration of GDCR computation at step $t$. In the thought stage, GDCR first identifies newly-cited entities $\Delta C_t$, namely entities that have appeared in previous observations but are explicitly cited for the first time in the current thought. Panels (a) and (b) show two such newly-cited entities: the entity in (a) has a smaller shortest-path distance to the answer node and therefore receives a larger graph-distance contribution score $c_q(v)$ than the entity in (b). In the observation stage, GDCR identifies newly-retrieved entities $\Delta O_t$ from the current search results. Panels (c) and (d) show two newly-retrieved entities, where the entity in (c) is closer to the answer node and receives a larger contribution score than the entity in (d).}
  \label{fig:gdcr}
\end{figure*}

We first define the training-time Entity-Relation (ER) graph, then introduce Graph-Distance Contribution Reward (GDCR) and Step Advantage Policy Optimization (SAPO).

\subsection{Training-Time ER Graph Construction}

As discussed in Section~\ref{sec:preliminaries}, the latent world graph $\mathcal{G}^{\mathrm{world}}$ is unobservable, and the latent task graph $\mathcal{G}_q$ for each query is therefore not directly available. During training, we construct a training-time ER graph $\hat{\mathcal{G}}_q=(\hat{\mathcal{V}}_q,\hat{\mathcal{E}}_q)$ for each query-answer pair, where nodes denote task-relevant entities, edges denote evidential relations, and the answer entity is annotated as the answer node $v_q^\ast$. Conceptually, the training-time ER graph $\hat{\mathcal{G}}_q$ serves as an observable proxy for the latent task graph.
$\hat{\mathcal{G}}_q$ helps the model learn to reason over entity relations and use search tools to find answer-relevant entities. The resulting $\hat{\mathcal{G}}_q$ serves as the graph basis for GDCR, which assigns step-level rewards according to progress toward the answer node. We apply structural checks and LLM-based verification during data synthesis; more details can be found in Appendix~\ref{data}.

\subsection{Graph-Distance Contribution Score}

Given the training-time ER graph $\hat{\mathcal{G}}_q$ and its answer node $v_q^\ast$, we estimate the contribution of each graph-linked entity by its shortest-path distance to the answer node. The intuition is that entities closer to the answer node indicate stronger progress toward the answer, and therefore provide a stronger step-level contribution signal.

Formally, for any entity node $v \in \hat{\mathcal{V}}_q$, let $\hat{\mathcal{P}}(v, v_q^\ast)$ denote the set of reachable paths from $v$ to the answer node $v_q^\ast$ in $\hat{\mathcal{G}}_q$. We define the shortest-path graph distance as:
\begin{equation}
    d_q(v, v_q^\ast) =
    \min_{p \in \hat{\mathcal{P}}(v, v_q^\ast)} |p| ,
\end{equation}
where $|p|$ is the number of edges in path $p$. Based on this distance, the graph-distance contribution score of node $v$ is defined as:
\begin{equation}
    c_q(v) = k^{-d_q(v, v_q^\ast)},
\end{equation}
where $k$ is a decay factor. In our experiments, we set $k=2$.

\subsection{Graph-Distance Contribution Reward}

Having established the training-time ER graph $\hat{\mathcal{G}}_q$ and the graph-distance contribution score $c_q(v)$ for each node, we define Graph-Distance Contribution Reward (GDCR) to quantify the extent to which a search agent discovers and explicitly cites relevant entities during a single step.
Let $\mathcal{H}_T$, as denoted in Eq.~\eqref{eq:Ht}, be a trajectory generated by the agent. We first define the following subsets of nodes:
\begin{equation}
    T_t \subseteq \hat{\mathcal{V}}_q, \quad O_t \subseteq \hat{\mathcal{V}}_q ,
\end{equation}
where $T_t$ represents the cumulative set of nodes explicitly cited in the thoughts up to step $t$ after they have appeared in observations; $O_t$ denotes the cumulative set of nodes appearing in the external observations up to step $t$. All sets are initialized as empty sets:
\begin{equation}
    T_0 = O_0 = \emptyset .
\end{equation}

Following the step structure in Section~\ref{sec:preliminaries}, GDCR computes rewards in two stages: it first scores newly-cited entities in the current thought, and then scores newly-retrieved entities in the current observation.

\noindent\textbf{Reward for Newly-Cited Entities.}
In the thought stage, the agent may explicitly cite entities that appeared in previous observations. To keep the reward step-level, we only count entities that are newly cited in the current thought:
\begin{equation}
    \Delta C_t = T_t \setminus T_{t-1}.
\end{equation}
The reward for newly-cited entities of step $t$ is defined as:
\begin{equation}
    r_t^{\mathrm{cite}} = \sum_{v \in \Delta C_t} c_q(v).
\end{equation}
Figures~\ref{fig:gdcr}(a) and \ref{fig:gdcr}(b) illustrate how graph distance distinguishes newly-cited entities.

\noindent\textbf{Reward for Newly-Retrieved Entities.}
In the observation stage, the action returns external observations that may contain entities not seen in previous observations. Let the newly-retrieved entities at step $t$ be denoted as:
\begin{equation}
    \Delta O_t = O_t \setminus O_{t-1}.
\end{equation}
The reward for newly-retrieved entities of step $t$ is then defined as:
\begin{equation}
    r_t^{\mathrm{ret}} = \sum_{v \in \Delta O_t} c_q(v).
\end{equation}
Figures~\ref{fig:gdcr}(c) and \ref{fig:gdcr}(d) illustrate the same graph-distance comparison for newly-retrieved entities.

Finally, GDCR for step $t$ is calculated as the sum of the reward for newly-cited entities and newly-retrieved entities:
\begin{equation}
    r_{t,g} = r_t^{\mathrm{cite}} + r_t^{\mathrm{ret}} .
\end{equation}

\subsection{Step Advantage Policy Optimization}

Step Advantage Policy Optimization (SAPO) is a general optimization method that converts process rewards into step-level advantages and combines them with trajectory-level outcome advantages. In this work, SAPO operationalizes the step-level credit derived from GDCR for Agentic Search training. For each query $q$, we sample a group of trajectories $\{\mathcal{H}^{(i)}\}_{i=1}^{G}$. Each trajectory receives a trajectory-level outcome reward according to final-answer correctness and format validity, which is then group-normalized to obtain the outcome advantage $\hat{A}^{(i)}_o$. This term anchors optimization to the final answer objective.

To introduce step-level credit assignment, SAPO uses the GDCR sequence $\mathbf{r}^{(i)}_g=(r^{(i)}_{1,g},\ldots,r^{(i)}_{T_i,g})$ computed along trajectory $\mathcal{H}^{(i)}$. Since raw GDCR scales vary across tasks and training-time ER graphs, we normalize the GDCR rewards within each trajectory and clip the resulting step-level signal:
\begin{equation}
    A^{(i)}_{t,g}
    =
    \mathrm{clip}\left(
    \frac{r^{(i)}_{t,g}-\mathrm{mean}_{t=1}^{T_i}(r^{(i)}_{t,g})}
    {\mathrm{std}_{t=1}^{T_i}(r^{(i)}_{t,g})+\varepsilon},
    -1, 1
    \right).
\end{equation}
The final advantage assigned to all tokens in step $t$ combines the outcome advantage and the GDCR-based step advantage:
\begin{equation}
    A^{(i)}_{t}
    =
    \hat{A}^{(i)}_o
    + \lambda |\hat{A}^{(i)}_o| A^{(i)}_{t,g},
    \quad t=1,\ldots,T_i ,
    \label{eq:lambda}
\end{equation}
where $\lambda$ controls the strength of the step-level signal. Thus, SAPO preserves the trajectory-level correctness objective while assigning different advantages to different steps according to GDCR.

Finally, the policy is optimized with the GRPO-style objective:
\begin{equation}
\small
\begin{aligned}
J(\theta)
=
&\mathbb{E}_{(q,y)\sim\mathcal{D},\,\mathcal{B}_q}
\Bigg[
\frac{1}{G}\sum_{i=1}^{G}
\frac{1}{|\mathcal{H}^{(i)}|}
\sum_{j=1}^{|\mathcal{H}^{(i)}|}
\\
&\min\!\Big(
\rho_{i,j}(\theta) A_{i,j}, \\
&\qquad
\mathrm{clip}(\rho_{i,j}(\theta),1-\epsilon_l,1+\epsilon_h) A_{i,j}
\Big)
\Bigg],
\end{aligned}
\end{equation}
where $\mathcal{B}_q=\{\mathcal{H}^{(i)}\}_{i=1}^{G}$ denotes the trajectory group, $\rho_{i,j}(\theta)$ is the token-level importance ratio, $A_{i,j}$ is the advantage of the step to which token $j$ belongs, and $\epsilon_l,\epsilon_h$ are clipping bounds.

\section{Experiment}

In this section, we evaluate SAPO through SFT and RL training, and compare it with outcome-reward training and existing search agents.

\subsection{Experiment Setup}

\noindent\textbf{Benchmarks.}
We evaluate our agent on four challenging deep research benchmarks: BrowseComp~\cite{wei2025browsecomp}, BrowseComp-ZH~\cite{zhou2025browsecompzh}, xbench-DS~\cite{chen2025xbench}, and GAIA~\cite{mialon2023gaia}. For GAIA, we use the text-only validation subset with 103 samples~\cite{li2025searcho1}. For the remaining benchmarks, we use the complete test sets.

\noindent\textbf{Metrics.}
Following prior work~\cite{team2025tongyi}, we adopt an LLM-as-a-judge evaluation protocol. Model answers are extracted from the final \textless answer\textgreater{} and \textless/answer\textgreater{} tags of the last response turn, and GPT-4o is used to judge correctness. Additional trajectory-format details can be found in Appendix~\ref{format}.

\noindent\textbf{Baselines.}
Our primary baseline is GRPO trained with outcome rewards and format penalties, using the same model architecture, training data, and context length as SAPO. We also include ARPO~\cite{dong2025agentic} as a branch-sampling policy optimization baseline at the 8B scale. Fully expanded tree-sampling methods are prohibitively expensive for Agentic Search training because each branch requires additional model generation and search interaction; therefore, we choose ARPO as a practical representative that samples branches according to a fixed strategy. For a controlled comparison, ARPO and SAPO use the same number of rollouts per prompt, which can be viewed as comparing the two methods under the same rollout budget. We also compare with representative open-source search agents under our evaluation environment and report publicly available results of strong closed-source systems as reference points. More experimental details can be found in Appendix~\ref{experiment}.

\noindent\textbf{Training Details.}
We conduct experiments with Qwen3-8B and Qwen3-30B-A3B-thinking-2507. The training pipeline contains a cold-start SFT phase followed by RL. For RL, we synthesize 1,000 graph-augmented QA pairs and 500 development examples for analysis. The 8B RL training uses group size 8 and a 32k context length; additional training and environment details can be found in Appendix~\ref{experiment}.

\subsection{RQ1: Is Graph Distance a Valid Progress Prior?}

We first validate whether moving closer to the answer node in the training-time ER graph is associated with final-answer correctness.

For each non-final step, we compute the shortest-path distance from the graph-linked entities newly retrieved or newly cited at that step to the answer node:
\begin{equation}
    d_t = \min_{v \in \Delta O_t \cup \Delta C_t} d_q(v, v_q^\ast).
\end{equation}
Here, \(\Delta O_t\) and \(\Delta C_t\) follow the definitions in Section~\ref{sec:methodology}. We then track the best distance reached up to step $t$:
\begin{equation}
    d^{\mathrm{best}}_t = \min_{s \le t} d_s .
\end{equation}
This prefix-based measure reflects the closest point reached so far.

\begin{table}[t]
\centering
\caption{History-best shortest-path distance to the answer node for correct and incorrect trajectories. Lower values indicate that the trajectory has reached entities closer to the answer node.}
\label{tab:rq1-best-distance}
\small
\setlength{\tabcolsep}{4pt}
\begin{adjustbox}{max width=.95\columnwidth}
\begin{tabular}{cccc}
\toprule
\textbf{Step} & \textbf{Correct} & \textbf{Incorrect} & \textbf{Diff.} \\
\midrule
1 & 1.836 & 2.819 & -0.983 \\
2 & 1.166 & 2.447 & -1.281 \\
3 & 1.288 & 2.569 & -1.280 \\
4 & 0.537 & 2.560 & -2.024 \\
5 & 0.590 & 2.995 & -2.405 \\
6 & 0.177 & 3.343 & -3.166 \\
7 & 0.137 & 4.193 & -4.057 \\
8 & 0.117 & 3.421 & -3.304 \\
\bottomrule
\end{tabular}%
\end{adjustbox}
\end{table}

Table~\ref{tab:rq1-best-distance} shows that correct trajectories consistently reach smaller history-best distances than incorrect ones, and the gap generally expands as search proceeds. As an auxiliary check, the GDCR step score is positively correlated with final-answer correctness ($r=0.334$, $p<10^{-179}$). These results support graph distance as a useful progress prior for GDCR.

\subsection{RQ2: Does SAPO Improve Search-Agent Performance?}

\begin{table*}[t]
\centering
\caption{Overall performance on four challenging deep search benchmarks. The best results in our environment are shown in bold, and the second-best results are underlined. Results of advanced models in their native environments are reported as reference points. ARPO is evaluated at the 8B scale with the same rollout budget per prompt as SAPO.}
\label{tab:main}
\small
\setlength{\tabcolsep}{4pt}
\begin{adjustbox}{max width=.95\textwidth}
\begin{tabular}{lccccc}
\toprule
\textbf{Model} & \textbf{Context Budget} & \textbf{BrowseComp-ZH} & \textbf{BrowseComp} & \textbf{xbench-DS} & \textbf{GAIA} \\
\midrule
\rowcolor{gray!30}
\multicolumn{6}{c}{\textit{\textbf{Advanced Models in Their Native Environments}}} \\
\midrule
GLM-4.6 \cite{zai2025glm46} & - & 45.1 & 49.5 & - & - \\
GLM-4.7 \cite{zai2025glm47} & - & 66.6 & 52.0 & - & - \\
OpenAI-o3 \cite{o3_o4_mini_openai} & - & 58.1 & 49.7 & 66.7 & 70.5 \\
Kimi K2 \cite{team2025kimi} & - & 28.8 & 14.1 & 50.0 & 57.7 \\
Web-30B-E-GRPO \cite{zhao2025repurposing} & - & 26.4 & 12.9 & 48.5 & 46.7 \\
\midrule
\rowcolor{gray!30}
\multicolumn{6}{c}{\textit{\textbf{Open-Source Search Agents in Our Environment}}} \\
\midrule
MiroThinker-v1.0-8B \cite{team2025mirothinker} & 64K & 16.3 & 2.2 & 13.3 & 31.2 \\
MiroThinker-v1.5-30B \cite{team2025mirothinker} & 128K & 5.7 & 0.7 & 5.0 & 23.3 \\
WebSailor-7B \cite{li2025websailor} & 32K & 9.8 & 1.0 & 9.3 & 35.9 \\
WebSailor-32B \cite{li2025websailor} & 32K & 18.0 & 5.5 & 11.0 & 63.1 \\
WebExplorer-8B \cite{liu2025webexplorer} & 64K & 6.2 & 2.9 & 2.0 & 50.5 \\
Tongyi DeepResearch \cite{team2025tongyi} & 128K & \underline{43.2} & \underline{40.7} & \underline{69.0} & \underline{68.9} \\
\midrule
\rowcolor{gray!30}
\multicolumn{6}{c}{\textit{\textbf{Our Agents}}} \\
\midrule
Qwen3-8B-SFT \cite{yang2025qwen3} & 64K & 17.9 & 2.7 & 18.0 & 38.8 \\
+GRPO \cite{shao2024deepseekmath} & 64K & 19.7 & 4.3 & 20.0 & 51.5 \\
+ARPO \cite{dong2025agentic} & 64K & 15.9 & 4.1 & 16.0 & 47.3 \\
+SAPO (ours) & 64K & 22.2 & 4.9 & 22.0 & 52.4 \\
\midrule
Qwen3-30B-A3B-thinking-SFT \cite{yang2025qwen3} & 128K & 27.3 & 13.9 & 53.0 & 57.3 \\
+GRPO \cite{shao2024deepseekmath} & 128K & 33.2 & 14.9 & 67.0 & 62.1 \\
+SAPO (ours) & 128K & \textbf{45.7} & \textbf{42.8} & \textbf{75.0} & \textbf{70.9} \\
\bottomrule
\end{tabular}%
\end{adjustbox}
\end{table*}

Table~\ref{tab:main} shows that SAPO consistently improves over SFT and GRPO across both model scales. At the 8B scale, SAPO also outperforms ARPO on all four benchmarks under the same rollout budget; the efficiency comparison in Appendix~\ref{app:efficiency} further suggests that ARPO incurs higher training cost. The gains are larger on Qwen3-30B-A3B-thinking, where SAPO achieves the best results among models evaluated in our environment. These results indicate that GDCR provides useful step-level credit beyond trajectory-level outcome rewards while preserving the final-answer objective.

\subsection{RQ3: Which Reward Components of GDCR Matter?}

We conduct an ablation study on Qwen3-8B to examine the contribution of the two reward components in GDCR.

\begin{table}[t]
\centering
\caption{Ablation study on the reward components in GDCR.}
\label{tab:reward-components}
\resizebox{\columnwidth}{!}{%
\begin{tabular}{lcccc}
\toprule
\textbf{Model} & \textbf{BC-ZH} & \textbf{BC} & \textbf{xbench-DS} & \textbf{GAIA} \\
\midrule
Qwen3-8B-SFT & 17.9 & 2.7 & 18.0 & 38.8 \\
+GRPO & 19.7 & 4.3 & 20.0 & 51.5 \\
+SAPO w/o newly-retrieved reward & 16.6 & 3.5 & 21.0 & 52.4 \\
+SAPO w/o newly-cited reward & 17.0 & 3.4 & 21.0 & 45.6 \\
+SAPO & \textbf{22.2} & \textbf{4.9} & \textbf{22.0} & \textbf{52.4} \\
\bottomrule
\end{tabular}%
}
\end{table}

Table~\ref{tab:reward-components} shows that both GDCR components are needed. Removing either newly-retrieved or newly-cited rewards hurts performance, while the full GDCR achieves the best or tied-best result on all four benchmarks.

\subsection{RQ4: How Robust and Efficient Is SAPO?}

SAPO provides step-level credit without generating additional search branches during training. On Qwen3-8B under the same hardware setting, SAPO takes 13.26 minutes per training step, adding only about 1\% overhead over GRPO, while ARPO costs 16.25 minutes per step, or 1.24$\times$ the GRPO cost, under the same rollout budget. More efficiency results, including the estimated cost of Tree-GRPO~\cite{ji2025tree}, can be found in Appendix~\ref{app:efficiency}.

\begin{table}[t]
\centering
\caption{Robustness to corrupted training-time ER graphs. In each noise column, the three numbers correspond to node deletion / noisy node injection / answer perturbation.}
\label{tab:graph-noise}
\resizebox{\columnwidth}{!}{%
\begin{tabular}{lcccc}
\toprule
\textbf{Dataset} & \textbf{GRPO} & \textbf{0\% Noise} & \textbf{10\% Noise} & \textbf{30\% Noise} \\
\midrule
xbench-DS & 20.0 & 22.0 & 22.0 / 21.0 / 20.0 & 22.0 / 22.0 / 18.0 \\
BC-ZH & 19.7 & 22.2 & 19.4 / 21.2 / 19.4 & 17.9 / 19.0 / 19.0 \\
\bottomrule
\end{tabular}%
}
\end{table}

Table~\ref{tab:graph-noise} shows that SAPO remains stable under moderate graph corruption. Performance drops more visibly under stronger noise, especially when the answer node is perturbed, confirming that graph quality remains important.

\subsection{RQ5: Does GDCR Improve Step-Level Search and Citation Behavior?}

\begin{figure}[t]
\centering
\includegraphics[width=\columnwidth]{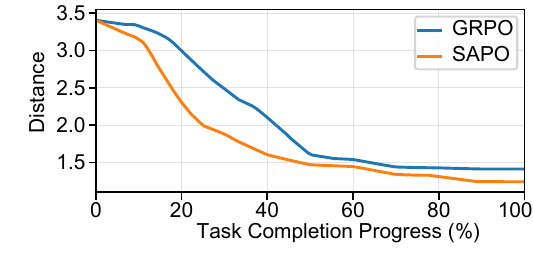}
\caption{Average shortest-path distance to the answer node as task progress increases. The comparison is conducted on the development set with Qwen3-8B models trained by GRPO and SAPO.}
\label{fig:step-dist}
\end{figure}

\begin{figure}[t]
\centering
\includegraphics[width=\columnwidth]{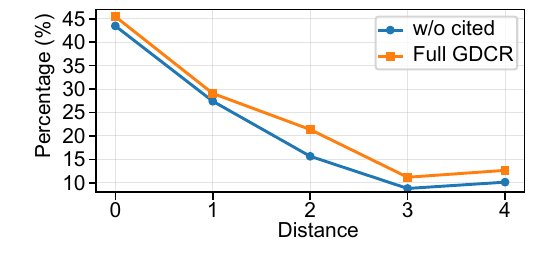}
\caption{Effectiveness analysis of newly-cited entities. The x-axis denotes the shortest-path distance to the answer node, and the y-axis denotes the percentage of retrieved entities that are explicitly cited in the trajectory.}
\label{fig:dist-percentage}
\end{figure}

\noindent\textbf{Progress Toward the Answer Node.}
We evaluate Qwen3-8B models trained with GRPO and SAPO on the development set. As shown in Figure~\ref{fig:step-dist}, SAPO yields a steeper decrease in average shortest-path distance and maintains a lower distance than GRPO throughout most of the trajectory.

\noindent\textbf{Newly-Cited Entities.}
Figure~\ref{fig:dist-percentage} shows that adding the newly-cited reward increases the proportion of retrieved entities that are later explicitly cited in the trajectory. This suggests that GDCR not only guides retrieval, but also encourages the agent to explicitly cite retrieved entities.

\subsection{Additional Analysis}

\noindent\textbf{Sensitivity to Step-Level Reward Weight.}
We first examine the sensitivity of SAPO to the step-level reward weight $\lambda$.

\begin{table}[t]
\centering
\caption{Ablation study on the step-level reward weight $\lambda$.}
\label{tab:lambda}
\resizebox{\columnwidth}{!}{%
\begin{tabular}{lcccc}
\toprule
\textbf{Model} & \textbf{BC-ZH} & \textbf{BC} & \textbf{xbench-DS} & \textbf{GAIA} \\
\midrule
Qwen3-8B-SFT & 17.9 & 2.7 & 18.0 & 38.8 \\
+SAPO ($\lambda=0.0$) & 19.7 & 4.3 & 20.0 & 51.5 \\
+SAPO ($\lambda=0.3$) & 20.8 & 3.5 & 21.0 & 47.6 \\
+SAPO ($\lambda=0.5$) & \textbf{22.2} & \textbf{4.9} & \textbf{22.0} & \textbf{52.4} \\
+SAPO ($\lambda=1.0$) & 20.4 & 2.9 & 20.0 & 45.3 \\
\bottomrule
\end{tabular}%
}
\end{table}

Table~\ref{tab:lambda} shows that removing the GDCR-based step advantage reduces SAPO to GRPO, while $\lambda=0.5$ gives the best overall performance.

\noindent\textbf{Sensitivity to Distance Decay.}
We then vary the graph-distance decay factor $k$ in the graph-distance contribution score.

\begin{table}[t]
\centering
\caption{Ablation study on the graph-distance decay factor $k$.}
\label{tab:decay}
\resizebox{\columnwidth}{!}{%
\begin{tabular}{lcccc}
\toprule
\textbf{Model} & \textbf{BC-ZH} & \textbf{BC} & \textbf{xbench-DS} & \textbf{GAIA} \\
\midrule
+SAPO ($k=1$) & 18.3 & 2.3 & \textbf{22.0} & 35.9 \\
+SAPO ($k=2$) & \textbf{22.2} & \textbf{4.9} & \textbf{22.0} & 52.4 \\
+SAPO ($k=4$) & 20.4 & 3.5 & 17.0 & \textbf{53.4} \\
\bottomrule
\end{tabular}%
}
\end{table}

Table~\ref{tab:decay} shows that $k=2$ gives the best overall performance, while $k=1$ removes the graph-distance distinction and degrades performance.

\noindent\textbf{Inference Budget Scaling.}
We also include inference budget scaling results under different context-length and tool-call limits in Appendix~\ref{app:scaling}.

\section{Related Work}
\label{sec:related work}

\noindent\textbf{Search Agent.} Search agents tackle complex, knowledge-intensive tasks that are difficult for single-pass inference by iteratively invoking external retrieval tools and reasoning over the returned evidence. Existing research in this domain primarily follows two technical trajectories: one line of work constructs large-scale, high-quality synthetic training data to improve complex information retrieval and reasoning capabilities \cite{li2025websailor2,wang2025explore,fang2025cognitive}; the other improves RL algorithms for search agents \cite{zhao2025repurposing,zhang2026chaining}. However, most RL methods still provide advantage signals at the trajectory level. Although some approaches employ tree sampling to allocate trajectory-level advantages to intermediate steps \cite{dong2025agentic,dong2025agenticE}, they remain constrained by high computational overhead, making it difficult to scale toward genuine step-level modeling.

\noindent\textbf{Process Reward in Agent Reinforcement Learning.} Process rewards improve agentic reinforcement learning by providing fine-grained credit assignment and more informative optimization signals. Existing research primarily constructs advantage signals at two granularities: one class combines process rewards with outcome rewards at the trajectory level \cite{tao2025webleaper,hao2025dynasearcher,zhao2025repurposing,deng2025atom,zhang2026chaining}, yet typically broadcasts the aggregated reward uniformly across all steps. The other class uses step-level advantage signals to guide model learning. Within this category, some methods introduce step-wise rewards based on PPO \cite{zheng2025stepsearch,ye2025process}, but depend on critic models; others perform dense tree sampling at each step, which becomes expensive for long-horizon multi-turn tasks \cite{wei2025reinforcingmultiturnreasoningllm}. Recent works therefore shift toward critical-step branch sampling strategies \cite{dong2025agentic,dong2025agenticE,ji2025tree,hou2025treerl,yang2025treerpo}, but still incur additional rollout cost. SAPO differs from these approaches by using GDCR as a training-time graph-distance reward, avoiding the overhead of critic models or dense tree searches while providing fine-grained step-level advantage assignment.

\section{Conclusion}

We presented GDCR, a graph-distance step-level reward for complex information-seeking tasks with definitive answers, and SAPO, an optimization method that converts process rewards into step-level advantages. By using a training-time ER graph to score newly-retrieved entities and newly-cited entities, GDCR provides step-level process rewards for Agentic Search training, while SAPO combines them with the trajectory-level correctness objective. Experiments and analyses show that this design improves Agentic Search training while keeping the overhead substantially lower than rollout-heavy step-level methods.

\section*{Limitations}

Our method is designed for complex information-seeking tasks with definitive answers. GDCR requires an answer node in the training-time ER graph, and therefore does not directly apply to open-ended generation, subjective analysis, or tasks without a clear endpoint.

The data synthesis and search environment also introduce reproducibility costs. The policy optimization objective and reward computation can be implemented independently of a specific retrieval backend, but absolute benchmark performance may still depend on the search engine, document index, and graph-construction pipeline used during training and evaluation.

\bibliography{custom}

\begin{thebibliography}{45}
\providecommand{\natexlab}[1]{#1}

\bibitem[{Chen et~al.(2025)Chen, Ren, Liu, Hu, Tian, Xie, Liu, Zhang, Liu, Gong et~al.}]{chen2025xbench}
Kaiyuan Chen, Yixin Ren, Yang Liu, Xiaobo Hu, Haotong Tian, Tianbao Xie, Fangfu Liu, Haoye Zhang, Hongzhang Liu, Yuan Gong, and 1 others. 2025.
\newblock xbench: Tracking agents productivity scaling with profession-aligned real-world evaluations.
\newblock \emph{arXiv preprint arXiv:2506.13651}.

\bibitem[{Deng et~al.(2025)Deng, Wang, Ying, Wu, Lin, Xiong, Dai, Yang, Zhang, Wang et~al.}]{deng2025atom}
Yong Deng, Guoqing Wang, Zhenzhe Ying, Xiaofeng Wu, Jinzhen Lin, Wenwen Xiong, Yuqin Dai, Shuo Yang, Zhanwei Zhang, Qiwen Wang, and 1 others. 2025.
\newblock Atom-searcher: Enhancing agentic deep research via fine-grained atomic thought reward.
\newblock \emph{arXiv preprint arXiv:2508.12800}.

\bibitem[{Dong et~al.(2025{\natexlab{a}})Dong, Bao, Wang, Zhao, Li, Jin, Yang, Mao, Zhang, Gai et~al.}]{dong2025agenticE}
Guanting Dong, Licheng Bao, Zhongyuan Wang, Kangzhi Zhao, Xiaoxi Li, Jiajie Jin, Jinghan Yang, Hangyu Mao, Fuzheng Zhang, Kun Gai, and 1 others. 2025{\natexlab{a}}.
\newblock Agentic entropy-balanced policy optimization.
\newblock \emph{arXiv preprint arXiv:2510.14545}.

\bibitem[{Dong et~al.(2025{\natexlab{b}})Dong, Mao, Ma, Bao, Chen, Wang, Chen, Du, Wang, Zhang et~al.}]{dong2025agentic}
Guanting Dong, Hangyu Mao, Kai Ma, Licheng Bao, Yifei Chen, Zhongyuan Wang, Zhongxia Chen, Jiazhen Du, Huiyang Wang, Fuzheng Zhang, and 1 others. 2025{\natexlab{b}}.
\newblock Agentic reinforced policy optimization.
\newblock \emph{arXiv preprint arXiv:2507.19849}.

\bibitem[{Fang et~al.(2025)Fang, Zhang, Wang, Wang, Qin, Wan, Ma, Zhang, Chen, Li et~al.}]{fang2025cognitive}
Tianqing Fang, Zhisong Zhang, Xiaoyang Wang, Rui Wang, Can Qin, Yuxuan Wan, Jun-Yu Ma, Ce~Zhang, Jiaqi Chen, Xiyun Li, and 1 others. 2025.
\newblock Cognitive kernel-pro: A framework for deep research agents and agent foundation models training.
\newblock \emph{arXiv preprint arXiv:2508.00414}.

\bibitem[{Gao et~al.(2025)Gao, Fu, Xie, Xu, He, Mei, Zhu, and Wu}]{gao2508beyond}
Jiaxuan Gao, Wei Fu, Minyang Xie, Shusheng Xu, Chuyi He, Zhiyu Mei, Banghua Zhu, and Yi~Wu. 2025.
\newblock Beyond ten turns: Unlocking long-horizon agentic search with large-scale asynchronous rl.
\newblock \emph{URL https://arxiv. org/abs/2508.07976}.

\bibitem[{Hao et~al.(2025)Hao, Feng, Zhang, and Wang}]{hao2025dynasearcher}
Chuzhan Hao, Wenfeng Feng, Yuewei Zhang, and Hao Wang. 2025.
\newblock Dynasearcher: Dynamic knowledge graph augmented search agent via multi-reward reinforcement learning.
\newblock \emph{arXiv preprint arXiv:2507.17365}.

\bibitem[{Hou et~al.(2025)Hou, Hu, Li, Lu, Tang, and Dong}]{hou2025treerl}
Zhenyu Hou, Ziniu Hu, Yujiang Li, Rui Lu, Jie Tang, and Yuxiao Dong. 2025.
\newblock Treerl: Llm reinforcement learning with on-policy tree search.
\newblock \emph{arXiv preprint arXiv:2506.11902}.

\bibitem[{Ji et~al.(2025)Ji, Ma, Wang, Chen, Chu, and Wu}]{ji2025tree}
Yuxiang Ji, Ziyu Ma, Yong Wang, Guanhua Chen, Xiangxiang Chu, and Liaoni Wu. 2025.
\newblock Tree search for llm agent reinforcement learning.
\newblock \emph{arXiv preprint arXiv:2509.21240}.

\bibitem[{Jin et~al.(2025)Jin, Zeng, Yue, Yoon, Arik, Wang, Zamani, and Han}]{jin2025search}
Bowen Jin, Hansi Zeng, Zhenrui Yue, Jinsung Yoon, Sercan Arik, Dong Wang, Hamed Zamani, and Jiawei Han. 2025.
\newblock Search-r1: Training llms to reason and leverage search engines with reinforcement learning.
\newblock \emph{arXiv preprint arXiv:2503.09516}.

\bibitem[{Li et~al.(2025{\natexlab{a}})Li, Zhang, Yin, Ye, Zhao, Zhang, Ou, Zhang, Wu, Wu et~al.}]{li2025websailor2}
Kuan Li, Zhongwang Zhang, Huifeng Yin, Rui Ye, Yida Zhao, Liwen Zhang, Litu Ou, Dingchu Zhang, Xixi Wu, Jialong Wu, and 1 others. 2025{\natexlab{a}}.
\newblock Websailor-v2: Bridging the chasm to proprietary agents via synthetic data and scalable reinforcement learning.
\newblock \emph{arXiv preprint arXiv:2509.13305}.

\bibitem[{Li et~al.(2025{\natexlab{b}})Li, Zhang, Yin, Zhang, Ou, Wu, Yin, Li, Tao, Wang et~al.}]{li2025websailor}
Kuan Li, Zhongwang Zhang, Huifeng Yin, Liwen Zhang, Litu Ou, Jialong Wu, Wenbiao Yin, Baixuan Li, Zhengwei Tao, Xinyu Wang, and 1 others. 2025{\natexlab{b}}.
\newblock Websailor: Navigating super-human reasoning for web agent.
\newblock \emph{arXiv preprint arXiv:2507.02592}.

\bibitem[{Li et~al.(2025{\natexlab{c}})Li, Dong, Jin, Zhang, Zhou, Zhu, Zhang, and Dou}]{li2025searcho1}
Xiaoxi Li, Guanting Dong, Jiajie Jin, Yuyao Zhang, Yujia Zhou, Yutao Zhu, Peitian Zhang, and Zhicheng Dou. 2025{\natexlab{c}}.
\newblock Search-o1: Agentic search-enhanced large reasoning models.
\newblock \emph{arXiv preprint arXiv:2501.05366}.

\bibitem[{Liu et~al.(2025)Liu, Li, Zhang, Li, Chen, Ji, Cheng, Wu, Du, Xu et~al.}]{liu2025webexplorer}
Junteng Liu, Yunji Li, Chi Zhang, Jingyang Li, Aili Chen, Ke~Ji, Weiyu Cheng, Zijia Wu, Chengyu Du, Qidi Xu, and 1 others. 2025.
\newblock Webexplorer: Explore and evolve for training long-horizon web agents.
\newblock \emph{arXiv preprint arXiv:2509.06501}.

\bibitem[{Ma et~al.(2025)Ma, Zhang, Zhao, Song, Wang, Sui, and Luo}]{ma2025stabilizing}
Wenhan Ma, Hailin Zhang, Liang Zhao, Yifan Song, Yudong Wang, Zhifang Sui, and Fuli Luo. 2025.
\newblock Stabilizing moe reinforcement learning by aligning training and inference routers.
\newblock \emph{arXiv preprint arXiv:2510.11370}.

\bibitem[{Mialon et~al.(2023)Mialon, Fourrier, Wolf, LeCun, and Scialom}]{mialon2023gaia}
Gr{\'e}goire Mialon, Cl{\'e}mentine Fourrier, Thomas Wolf, Yann LeCun, and Thomas Scialom. 2023.
\newblock Gaia: a benchmark for general ai assistants.
\newblock In \emph{The Twelfth International Conference on Learning Representations}.

\bibitem[{{Moonshot AI}(2025)}]{KimiResearcher2025}
{Moonshot AI}. 2025.
\newblock \href {https://moonshotai.github.io/Kimi-Researcher/} {Kimi-researcher end-to-end rl training for emerging agentic capabilities}.

\bibitem[{{OpenAI}(2025{\natexlab{a}})}]{o3_o4_mini_openai}
{OpenAI}. 2025{\natexlab{a}}.
\newblock \href {https://openai.com/zh-Hans-CN/index/introducing-o3-and-o4-mini/} {Introducing openai o3 and o4-mini}.

\bibitem[{{OpenAI}(2025{\natexlab{b}})}]{OpenAIDeepResearch2025}
{OpenAI}. 2025{\natexlab{b}}.
\newblock \href {https://openai.com/zh-Hans-CN/index/introducing-deep-research/} {Openai deep research}.

\bibitem[{{Quark}(2024)}]{quark2024search}
{Quark}. 2024.
\newblock Quark ai business search api.
\newblock \url{https://vt.quark.cn/blm/qk-ai-business-page-915/index?x_render_type=stream_ssr}.
\newblock Accessed: 2024.

\bibitem[{Shao et~al.(2024)Shao, Wang, Zhu, Xu, Song, Bi, Zhang, Zhang, Li, Wu et~al.}]{shao2024deepseekmath}
Zhihong Shao, Peiyi Wang, Qihao Zhu, Runxin Xu, Junxiao Song, Xiao Bi, Haowei Zhang, Mingchuan Zhang, YK~Li, Yang Wu, and 1 others. 2024.
\newblock Deepseekmath: Pushing the limits of mathematical reasoning in open language models.
\newblock \emph{arXiv preprint arXiv:2402.03300}.

\bibitem[{Shinn et~al.(2023)Shinn, Cassano, Gopinath, Narasimhan, and Yao}]{shinn2023reflexion}
Noah Shinn, Federico Cassano, Ashwin Gopinath, Karthik Narasimhan, and Shunyu Yao. 2023.
\newblock Reflexion: Language agents with verbal reinforcement learning.
\newblock \emph{Advances in Neural Information Processing Systems}, 36:8634--8652.

\bibitem[{Song et~al.(2025)Song, Jiang, Min, Chen, Chen, Zhao, Fang, and Wen}]{song2025r1}
Huatong Song, Jinhao Jiang, Yingqian Min, Jie Chen, Zhipeng Chen, Wayne~Xin Zhao, Lei Fang, and Ji-Rong Wen. 2025.
\newblock R1-searcher: Incentivizing the search capability in llms via reinforcement learning.
\newblock \emph{arXiv preprint arXiv:2503.05592}.

\bibitem[{Tao et~al.(2025)Tao, Shen, Li, Yin, Wu, Li, Zhang, Yin, Ye, Zhang et~al.}]{tao2025webleaper}
Zhengwei Tao, Haiyang Shen, Baixuan Li, Wenbiao Yin, Jialong Wu, Kuan Li, Zhongwang Zhang, Huifeng Yin, Rui Ye, Liwen Zhang, and 1 others. 2025.
\newblock Webleaper: Empowering efficiency and efficacy in webagent via enabling info-rich seeking.
\newblock \emph{arXiv preprint arXiv:2510.24697}.

\bibitem[{Team et~al.(2025{\natexlab{a}})Team, Bai, Bao, Chen, Chen, Chen, Chen, Chen, Chen, Chen et~al.}]{team2025kimi}
Kimi Team, Yifan Bai, Yiping Bao, Guanduo Chen, Jiahao Chen, Ningxin Chen, Ruijue Chen, Yanru Chen, Yuankun Chen, Yutian Chen, and 1 others. 2025{\natexlab{a}}.
\newblock Kimi k2: Open agentic intelligence.
\newblock \emph{arXiv preprint arXiv:2507.20534}.

\bibitem[{Team et~al.(2025{\natexlab{b}})Team, Bai, Bing, Chen, Chen, Chen, Chen, Chen, Dai, Dong et~al.}]{team2025mirothinker}
MiroMind Team, Song Bai, Lidong Bing, Carson Chen, Guanzheng Chen, Yuntao Chen, Zhe Chen, Ziyi Chen, Jifeng Dai, Xuan Dong, and 1 others. 2025{\natexlab{b}}.
\newblock Mirothinker: Pushing the performance boundaries of open-source research agents via model, context, and interactive scaling.
\newblock \emph{arXiv preprint arXiv:2511.11793}.

\bibitem[{Team et~al.(2025{\natexlab{c}})Team, Li, Zhang, Zhang, Huang, Li, Chen, Yin, Wu, Zhou et~al.}]{team2025tongyi}
Tongyi~DeepResearch Team, Baixuan Li, Bo~Zhang, Dingchu Zhang, Fei Huang, Guangyu Li, Guoxin Chen, Huifeng Yin, Jialong Wu, Jingren Zhou, and 1 others. 2025{\natexlab{c}}.
\newblock Tongyi deepresearch technical report.
\newblock \emph{arXiv preprint arXiv:2510.24701}.

\bibitem[{Wang et~al.(2025)Wang, Zhang, Ma, Zhang, Wang, Chen, Xue, Fang, Zhang, Zhang et~al.}]{wang2025explore}
Rui Wang, Ce~Zhang, Jun-Yu Ma, Jianshu Zhang, Hongru Wang, Yi~Chen, Boyang Xue, Tianqing Fang, Zhisong Zhang, Hongming Zhang, and 1 others. 2025.
\newblock Explore to evolve: Scaling evolved aggregation logic via proactive online exploration for deep research agents.
\newblock \emph{arXiv preprint arXiv:2510.14438}.

\bibitem[{Wei et~al.(2025{\natexlab{a}})Wei, Sun, Papay, McKinney, Han, Fulford, Chung, Passos, Fedus, and Glaese}]{wei2025browsecomp}
Jason Wei, Zhiqing Sun, Spencer Papay, Scott McKinney, Jeffrey Han, Isa Fulford, Hyung~Won Chung, Alex~Tachard Passos, William Fedus, and Amelia Glaese. 2025{\natexlab{a}}.
\newblock Browsecomp: A simple yet challenging benchmark for browsing agents.
\newblock \emph{arXiv preprint arXiv:2504.12516}.

\bibitem[{Wei et~al.(2022)Wei, Wang, Schuurmans, Bosma, Xia, Chi, Le, Zhou et~al.}]{wei2022chain}
Jason Wei, Xuezhi Wang, Dale Schuurmans, Maarten Bosma, Fei Xia, Ed~Chi, Quoc~V Le, Denny Zhou, and 1 others. 2022.
\newblock Chain-of-thought prompting elicits reasoning in large language models.
\newblock \emph{Advances in neural information processing systems}, 35:24824--24837.

\bibitem[{Wei et~al.(2025{\natexlab{b}})Wei, Zeng, Li, Brown, Frunza, Deng, Schneider, Nevmyvaka, Zhao, Garcia, and Hong}]{wei2025reinforcingmultiturnreasoningllm}
Quan Wei, Siliang Zeng, Chenliang Li, William Brown, Oana Frunza, Wei Deng, Anderson Schneider, Yuriy Nevmyvaka, Yang~Katie Zhao, Alfredo Garcia, and Mingyi Hong. 2025{\natexlab{b}}.
\newblock \href {https://arxiv.org/abs/2505.11821} {Reinforcing multi-turn reasoning in llm agents via turn-level reward design}.
\newblock \emph{Preprint}, arXiv:2505.11821.

\bibitem[{Wu et~al.(2025)Wu, Li, Fang, Yin, Zhang, Tao, Zhang, Xi, Fu, Jiang et~al.}]{wu2025webdancer}
Jialong Wu, Baixuan Li, Runnan Fang, Wenbiao Yin, Liwen Zhang, Zhengwei Tao, Dingchu Zhang, Zekun Xi, Gang Fu, Yong Jiang, and 1 others. 2025.
\newblock Webdancer: Towards autonomous information seeking agency.
\newblock \emph{arXiv preprint arXiv:2505.22648}.

\bibitem[{Yang et~al.(2025{\natexlab{a}})Yang, Li, Yang, Zhang, Hui, Zheng, Yu, Gao, Huang, Lv et~al.}]{yang2025qwen3}
An~Yang, Anfeng Li, Baosong Yang, Beichen Zhang, Binyuan Hui, Bo~Zheng, Bowen Yu, Chang Gao, Chengen Huang, Chenxu Lv, and 1 others. 2025{\natexlab{a}}.
\newblock Qwen3 technical report.
\newblock \emph{arXiv preprint arXiv:2505.09388}.

\bibitem[{Yang et~al.(2025{\natexlab{b}})Yang, Guo, Huang, Liang, Wang, and Tang}]{yang2025treerpo}
Zhicheng Yang, Zhijiang Guo, Yinya Huang, Xiaodan Liang, Yiwei Wang, and Jing Tang. 2025{\natexlab{b}}.
\newblock Treerpo: Tree relative policy optimization.
\newblock \emph{arXiv preprint arXiv:2506.05183}.

\bibitem[{Yao et~al.(2022)Yao, Zhao, Yu, Du, Shafran, Narasimhan, and Cao}]{yao2022react}
Shunyu Yao, Jeffrey Zhao, Dian Yu, Nan Du, Izhak Shafran, Karthik~R Narasimhan, and Yuan Cao. 2022.
\newblock React: Synergizing reasoning and acting in language models.
\newblock In \emph{The eleventh international conference on learning representations}.

\bibitem[{Ye et~al.(2025)Ye, Zhang, Jiang, and Huang}]{ye2025process}
Yufan Ye, Ting Zhang, Wenbin Jiang, and Hua Huang. 2025.
\newblock Process-supervised reinforcement learning for code generation.
\newblock \emph{arXiv preprint arXiv:2502.01715}.

\bibitem[{Yu et~al.(2025)Yu, Zhang, Zhu, Yuan, Zuo, Yue, Dai, Fan, Liu, Liu et~al.}]{yu2025dapo}
Qiying Yu, Zheng Zhang, Ruofei Zhu, Yufeng Yuan, Xiaochen Zuo, Yu~Yue, Weinan Dai, Tiantian Fan, Gaohong Liu, Lingjun Liu, and 1 others. 2025.
\newblock Dapo: An open-source llm reinforcement learning system at scale.
\newblock \emph{arXiv preprint arXiv:2503.14476}.

\bibitem[{Z.ai(2025{\natexlab{a}})}]{zai2025glm46}
Z.ai. 2025{\natexlab{a}}.
\newblock \href {https://z.ai/blog/glm-4.6} {Glm-4.6: Advanced agentic, reasoning and coding capabilities}.

\bibitem[{Z.ai(2025{\natexlab{b}})}]{zai2025glm47}
Z.ai. 2025{\natexlab{b}}.
\newblock \href {https://z.ai/blog/glm-4.7} {Glm-4.7: Advancing the coding capability}.

\bibitem[{Zhang et~al.(2026)Zhang, Lv, Feng, Hou, and Li}]{zhang2026chaining}
Jiajie Zhang, Xin Lv, Ling Feng, Lei Hou, and Juanzi Li. 2026.
\newblock Chaining the evidence: Robust reinforcement learning for deep search agents with citation-aware rubric rewards.
\newblock \emph{arXiv preprint arXiv:2601.06021}.

\bibitem[{Zhao et~al.(2025)Zhao, Li, Wu, Zhang, Zhang, Li, Song, Chen, Wang, Wang et~al.}]{zhao2025repurposing}
Yida Zhao, Kuan Li, Xixi Wu, Liwen Zhang, Dingchu Zhang, Baixuan Li, Maojia Song, Zhuo Chen, Chenxi Wang, Xinyu Wang, and 1 others. 2025.
\newblock Repurposing synthetic data for fine-grained search agent supervision.
\newblock \emph{arXiv preprint arXiv:2510.24694}.

\bibitem[{Zheng et~al.(2025{\natexlab{a}})Zheng, Liu, Li, Chen, Yu, Gao, Dang, Liu, Men, Yang et~al.}]{zheng2025group}
Chujie Zheng, Shixuan Liu, Mingze Li, Xiong-Hui Chen, Bowen Yu, Chang Gao, Kai Dang, Yuqiong Liu, Rui Men, An~Yang, and 1 others. 2025{\natexlab{a}}.
\newblock Group sequence policy optimization.
\newblock \emph{arXiv preprint arXiv:2507.18071}.

\bibitem[{Zheng et~al.(2025{\natexlab{b}})Zheng, An, Wang, Wang, and Wu}]{zheng2025stepsearch}
Xuhui Zheng, Kang An, Ziliang Wang, Yuhang Wang, and Yichao Wu. 2025{\natexlab{b}}.
\newblock Stepsearch: Igniting llms search ability via step-wise proximal policy optimization.
\newblock In \emph{Proceedings of the 2025 Conference on Empirical Methods in Natural Language Processing}, pages 21816--21841.

\bibitem[{Zheng et~al.(2025{\natexlab{c}})Zheng, Fu, Hu, Cai, Ye, Lu, and Liu}]{zheng2025deepresearcher}
Yuxiang Zheng, Dayuan Fu, Xiangkun Hu, Xiaojie Cai, Lyumanshan Ye, Pengrui Lu, and Pengfei Liu. 2025{\natexlab{c}}.
\newblock Deepresearcher: Scaling deep research via reinforcement learning in real-world environments.
\newblock \emph{arXiv preprint arXiv:2504.03160}.

\bibitem[{Zhou et~al.(2025)Zhou, Leon, Ying, Zhang, Shao, Ye, Chong, Jin, Xie, Cao et~al.}]{zhou2025browsecompzh}
Peilin Zhou, Bruce Leon, Xiang Ying, Can Zhang, Yifan Shao, Qichen Ye, Dading Chong, Zhiling Jin, Chenxuan Xie, Meng Cao, and 1 others. 2025.
\newblock Browsecomp-zh: Benchmarking web browsing ability of large language models in chinese.
\newblock \emph{arXiv preprint arXiv:2504.19314}.

\end{thebibliography}

\clearpage
\appendix
\section{Format}
\label{format}

Our ReAct framework follows the format shown in Figure~\ref{fig:trajectory_format}:

\begin{figure}[htbp]
    \centering
    \begin{tcolorbox}[
        title={Trajectory Format},
        colframe=darkgray,
        colback=white,
        coltitle=white,
        fonttitle=\large\bfseries,
        arc=1mm,
        boxrule=0.5mm,
        left=8pt, right=8pt, top=8pt, bottom=8pt,
        fontupper=\ttfamily\footnotesize
    ]
    
        \textcolor{cyan}{<|im\_start|>user} \\
        question here \\
        \textcolor{cyan}{<|im\_end|>} \\
        \textcolor{blue}{<|im\_start|>assistant} \\
        \textcolor{blue}{<think>} \\
        thinking process here \\
        \textcolor{blue}{</think>} \\
        \textcolor{blue}{<tool\_call>} \\
        \{"name": "tool name", "arguments": \{ ... \}\} \\
        \textcolor{blue}{</tool\_call>} \textcolor{blue}{<|im\_end|>} \\
        \textcolor{orange}{<|im\_start|>user} \\
        \textcolor{orange}{<tool\_response>} \\
        tool response here \\
        \textcolor{orange}{</tool\_response>} \textcolor{orange}{<|im\_end|>} \\
        (more thinking processes, tool calls and tool responses...) \\
        \textcolor{blue}{<|im\_start|>assistant} \\
        \textcolor{blue}{<think>} \\
        thinking process here \\
        \textcolor{blue}{</think>} \\
        \textcolor{red}{<answer>} \\
        answer here \\
        \textcolor{red}{</answer>} \textcolor{blue}{<|im\_end|>}
    
    \end{tcolorbox}
    \caption{Format of search agent trajectory.}
    \label{fig:trajectory_format}
\end{figure}

\section{QA Pair Synthesis with Graphs}
\label{data}

To facilitate the calculation of GDCR, it is imperative that each QA pair be associated with a training-time ER graph encompassing relevant entities and relationships. Since existing open-source data synthesis workflows fail to meet this requirement, we established two distinct data synthesis pipelines. Specifically, we synthesize data employing the following two methodologies:

\noindent\textbf{Knowledge Graph-Centric Approach.} For the Chinese dataset, we constructed a large-scale knowledge graph utilizing page content from Baidu Baike. We initiate the process by selecting an arbitrary starting node within the knowledge graph and employing a random walk strategy to derive a candidate training-time Entity-Relation (ER) graph. To ensure graph quality and reduce shortcut solutions, we apply three validations during synthesis. First, \textit{distant unreachability} checks that nodes multiple hops away from the answer cannot directly retrieve the answer. Second, \textit{intermediate unskippability} checks that target nodes cannot be retrieved by skipping their immediate predecessors along the path. Third, \textit{connectivity} checks that adjacent nodes can retrieve each other, ensuring valid local paths. We then task GPT-4o with generating initial questions based on the validated training-time ER graph, which are further complicated through techniques such as obfuscation. After QA generation, we apply two additional quality-control steps: filtering out trivial questions that can be directly answered by the model, and using LLM-as-a-judge to verify logical consistency and the correspondence among the question, answer, and training-time ER graph.

\noindent\textbf{Search Agent-Centric Approach.} For the English dataset, we leverage the open-source data synthesis pipeline from Asearcher \cite{gao2508beyond}. We keep the original QA synthesis process and add an ER graph construction step: after the final questions are generated, we instruct the model to construct a corresponding training-time ER graph for each QA pair based on the Supporting Statements. The specific prompt utilized for this process is illustrated in Figure~\ref{fig:prompt_graph}.

\begin{tcolorbox}[
    title={Prompt for Generating ER Graph},
    breakable,
    colframe=darkgray,
    colback=white,
    coltitle=white,
    fonttitle=\normalsize\bfseries,
    arc=1mm,
    boxrule=0.5mm,
    left=4pt, right=4pt, top=4pt, bottom=4pt,
    fontupper=\ttfamily\scriptsize
]
        
    \setlength{\parskip}{0.25em}

    You are a knowledge graph extraction agent.

    Given a Question–Answer (QA) pair and supporting factual statements, extract all factual triples required to represent the complete semantics of the QA pair.

    \textbf{\#\# Input}

    Question:\\
    \{question\}

    Answer:\\
    \{answer\}

    Supporting Statements:\\
    \{statements\}

    \textbf{\#\# Requirements}

    Extract factual triples in the form (subject, predicate, object).

    The extracted triples must collectively cover:
    \begin{itemize}[leftmargin=*, nosep]
        \item facts implied by the question,
        \item facts explicitly asserted by the answer,
        \item missing or implicit information resolved using the supporting statements.
    \end{itemize}

    \textbf{\#\#\# Mandatory Answer Inclusion}

    The answer "\{answer\}" must appear exactly as given as an entity in at least one triple.
    This requirement overrides the entity length constraint: the answer entity may exceed 3 words and must not be shortened, paraphrased, or modified.

    \textbf{\#\#\# Extraction Guidelines}

    \begin{itemize}[leftmargin=*, nosep]
        \item Extract only explicitly stated facts.
        \item Do not infer or introduce new information.
        \item Do not include questions, reasoning steps, or explanations.
        \item Each triple must express a single atomic fact.
        \item Avoid pronouns and vague references.
    \end{itemize}

    \textbf{\#\#\# Format Requirements}

    1. Triple Structure
    All triples must follow the string-based format [entity, relation, entity].
    
    2. Entity Length Constraint
    All entities must be $\le$ 3 words, except the answer entity which must appear exactly as given.
    If a non-answer entity exceeds 3 words:
    \begin{itemize}[leftmargin=*, nosep]
        \item retain only the essential identifying words, or
        \item split the information into multiple triples using intermediate entities.
    \end{itemize}

    \textbf{\#\#\# Pre-Output Validation}

    Before producing the final output, verify that:
    
    1. The answer appears exactly as given in at least one triple.
    
    2. Every non-answer entity contains no more than 3 words.
    
    3. All triples are valid 3-element string lists.
    
    4. Relations are concise and semantically clear.

    \textbf{\#\# Output Format}

    Return a JSON-like array of triples, formatted as:
    
    [\\
    \hspace*{1em} ["entity1", "relation", "entity2"],\\
    \hspace*{1em} ["entity1", "relation", "entity2"]\\
    ]

\end{tcolorbox}
\captionof{figure}{Prompt for generating ER Graph.}
\label{fig:prompt_graph}

\section{Experiment Detail}
\label{experiment}

\subsection{Evaluation Baselines.}

Regarding open-source models, we evaluated Mirothinker-8B/30B \cite{team2025mirothinker}, WebSailor-7B/32B \cite{li2025websailor}, WebExplorer-8B \cite{liu2025webexplorer}, and Tongyi DeepResearch \cite{team2025tongyi} under our current experimental conditions. Specifically, we assessed the 30B models using a 128K context budget and the 7B/8B models using a 64K context budget. An exception was made for WebSailor-7B/32B, which was tested at its maximum position embedding limit of 32K. For closed-source models (GLM-4.6 \cite{zai2025glm46}, GLM-4.7 \cite{zai2025glm47}, OpenAI-o3 \cite{o3_o4_mini_openai}, Kimi K2 \cite{team2025kimi}, and Web-30B-E-GRPO \cite{zhao2025repurposing}), we report scores derived from official technical reports or prior papers.

\subsection{Search Environment.}

We employed an internal KG-search tool as our primary search interface, which integrates Quark's third-party search API~\cite{quark2024search} and leverages approximately 8 billion documents from our proprietary knowledge base for document retrieval. The system performs unified ranking and returns the top 20 search results, with each result providing the most relevant passage from the full text, limited to 300 characters per passage. During the training phase of the 8B model, we used only the top 5 retrieved results. Although our production environment also supports PDF and document parsing capabilities, as well as a visit tool for retrieving complete webpage content, these additional tools were excluded from our experimental comparisons to ensure fair evaluation across all models.

\subsection{Training Details.}
\noindent\textbf{Qwen3-8B SFT.} For the cold-start SFT phase on the 8B model, we curate 492 high-quality trajectories with a maximum length of 32k tokens. These trajectories are collected through rejection sampling with Qwen3-235B-A22B on our synthesized dataset. We fine-tune the model for 5 epochs with a batch size of 16 and a learning rate of $4 \times 10^{-5}$.

\noindent\textbf{Qwen3-8B RL.} For RL, we synthesize 1,000 QA pairs with training-time ER graphs, including 600 Chinese and 400 English samples, and use 500 examples as a development set for analysis. The 8B RL training uses a 32k context length, a tool-call budget of 20, a training batch size of 32, a PPO batch size of 32, a group size of 8, and a learning rate of $2 \times 10^{-6}$ for approximately 3 epochs. This training run is conducted for 20 hours on 16 H200 141G GPUs.

\noindent\textbf{Qwen3-30B-A3B SFT.} For the 30B model, due to differences in context length requirements and MoE architecture demands for training data, we utilize 1,400 high-quality QA pairs generated using the same methodology as the 8B model. Through rejection sampling with Qwen3-235B, we obtain approximately 10,000 training trajectories. The 30B model is trained with a learning rate of $7 \times 10^{-6}$, batch size of 512, for 3 epochs, with a maximum context length of 64k tokens, warmup ratio of 0.03, and moe-aux-loss-coeff of 0.02.

\noindent\textbf{Qwen3-30B-A3B RL.} The 30B model employs similar RL settings as the 8B model, but with key modifications for MoE training stability. We incorporate the router replay method~\cite{ma2025stabilizing} to address the training-inference consistency issues in MoE models, where routing mechanisms can introduce instability and lead to catastrophic RL training collapse. This method records routing distributions from the inference engine and replays them during training to mitigate the discrepancy in routing behaviors. Additionally, leveraging the A3B model's enhanced context capabilities, we use a 64k context limit (24k for prompts + 40k for responses) during RL training. Due to resource constraints, all reported test results are from single runs.

\subsection{Training Efficiency}
\label{app:efficiency}

\begin{table}[!t]
\centering
\caption{Training efficiency comparison. Time is measured as total optimization time per training step on Qwen3-8B. The Tree-GRPO cost is an estimate based on complete tree expansion rather than a measured training run.}
\label{tab:efficiency}
\small
\setlength{\tabcolsep}{4pt}
\begin{adjustbox}{max width=\columnwidth}
\begin{tabular}{lcc}
\toprule
\textbf{Method} & \textbf{Time/Step (min)} & \textbf{Relative Cost} \\
\midrule
GRPO & 13.13 & 1.00$\times$ \\
SAPO (ours) & 13.26 & 1.01$\times$ \\
ARPO~\cite{dong2025agentic} & 16.25 & 1.24$\times$ \\
Tree-GRPO~\cite{ji2025tree} ($B=2$, estimated) & $\sim$2107 & $\sim$160.47$\times$ \\
\bottomrule
\end{tabular}%
\end{adjustbox}
\end{table}

As shown in Table~\ref{tab:efficiency}, SAPO preserves the training cost profile of linear trajectory optimization, while ARPO incurs additional branch-sampling cost and complete tree expansion becomes substantially more expensive for long-horizon search trajectories.

\subsection{Inference Budget Scaling}
\label{app:scaling}

We further evaluate how the 30B model scales under different inference budgets. These experiments vary the context-length limit and the tool-call limit on BrowseComp-ZH and BrowseComp. The results are intended as supplementary evidence for inference-time efficiency and are not part of the main controlled training-cost comparison in Table~\ref{tab:efficiency}.

\begingroup
\setlength{\abovecaptionskip}{4pt}
\setlength{\belowcaptionskip}{4pt}

\begin{table}[!t]
\centering
\caption{Accuracy under different context-length limits.}
\label{tab:context-scaling}
\small
\setlength{\tabcolsep}{4pt}
\begin{adjustbox}{max width=\columnwidth}
\begin{tabular}{lcccccccc}
\toprule
\multirow{2}{*}{\textbf{Model}} & \multicolumn{4}{c}{\textbf{BrowseComp-ZH}} & \multicolumn{4}{c}{\textbf{BrowseComp}} \\
\cmidrule(lr){2-5}\cmidrule(lr){6-9}
 & \textbf{16K} & \textbf{32K} & \textbf{64K} & \textbf{128K} & \textbf{16K} & \textbf{32K} & \textbf{64K} & \textbf{128K} \\
\midrule
GLM-4.6 & \underline{33.6} & \underline{36.7} & \textbf{40.5} & \underline{45.1} & \textbf{35.8} & 34.8 & \textbf{44.3} & \textbf{49.5} \\
Tongyi-30B & 32.5 & \underline{36.7} & 38.8 & 43.2 & 29.8 & \underline{35.2} & 38.5 & 40.7 \\
SAPO-30B & \textbf{33.9} & \textbf{37.4} & \underline{39.8} & \textbf{45.7} & \underline{30.5} & \textbf{35.9} & 37.3 & \underline{42.8} \\
GRPO-30B & 26.9 & 28.7 & 33.9 & 33.2 & 23.9 & 27.9 & 34.2 & 14.9 \\
Miro-30B & 27.3 & 32.5 & 36.3 & 5.7 & 28.8 & 31.7 & \underline{39.1} & 0.7 \\
\bottomrule
\end{tabular}%
\end{adjustbox}
\end{table}

\begin{table}[!t]
\centering
\caption{Token consumption at the 128K context-length limit. Avg, Min, and Max are reported in thousands of tokens.}
\label{tab:token-consumption}
\small
\setlength{\tabcolsep}{4pt}
\begin{adjustbox}{max width=\columnwidth}
\begin{tabular}{lcccccccc}
\toprule
\multirow{2}{*}{\textbf{Model}} & \multicolumn{4}{c}{\textbf{BrowseComp-ZH}} & \multicolumn{4}{c}{\textbf{BrowseComp}} \\
\cmidrule(lr){2-5}\cmidrule(lr){6-9}
 & \textbf{Acc.} & \textbf{Avg} & \textbf{Min} & \textbf{Max} & \textbf{Acc.} & \textbf{Avg} & \textbf{Min} & \textbf{Max} \\
\midrule
GLM-4.6 & \underline{45.1} & 40.6 & 15.4 & \textbf{94.5} & \textbf{49.5} & 37.8 & 19.7 & \underline{53.6} \\
Tongyi-30B & 43.2 & \underline{37.8} & \underline{4.7} & 127.8 & 40.7 & \textbf{27.3} & 24.4 & \textbf{31.9} \\
SAPO-30B & \textbf{45.7} & \textbf{24.9} & 8.3 & \underline{102.9} & \underline{42.8} & \underline{29.3} & \underline{13.5} & 94.5 \\
GRPO-30B & 33.2 & 48.3 & \textbf{2.6} & 119.2 & 14.9 & 38.2 & \textbf{9.2} & 103.7 \\
Miro-30B & 5.7 & 67.9 & 47.1 & 128.0 & 0.7 & 87.4 & 62.5 & 128.0 \\
\bottomrule
\end{tabular}%
\end{adjustbox}
\end{table}

\begin{table}[!t]
\centering
\caption{Accuracy under different tool-call limits.}
\label{tab:tool-call-scaling}
\small
\setlength{\tabcolsep}{4pt}
\begin{adjustbox}{max width=\columnwidth}
\begin{tabular}{lcccccccc}
\toprule
\multirow{2}{*}{\textbf{Model}} & \multicolumn{4}{c}{\textbf{BrowseComp-ZH}} & \multicolumn{4}{c}{\textbf{BrowseComp}} \\
\cmidrule(lr){2-5}\cmidrule(lr){6-9}
 & \textbf{4} & \textbf{16} & \textbf{64} & \textbf{256} & \textbf{4} & \textbf{16} & \textbf{64} & \textbf{256} \\
\midrule
GLM-4.6 & \textbf{35.9} & \underline{42.1} & \textbf{49.2} & \underline{45.1} & \textbf{34.3} & \textbf{44.4} & \textbf{49.4} & \textbf{49.5} \\
Tongyi-30B & \underline{33.9} & 40.1 & 45.3 & 43.2 & \underline{28.4} & 35.6 & 39.8 & 40.7 \\
SAPO-30B & 28.7 & \textbf{43.2} & \underline{47.4} & \textbf{45.7} & 24.7 & \underline{36.8} & \underline{43.6} & \underline{42.8} \\
GRPO-30B & 26.9 & 36.3 & 33.2 & 33.2 & 11.2 & 17.4 & 15.6 & 14.9 \\
Miro-30B & 31.5 & 37.0 & 18.7 & 5.7 & 18.7 & 31.2 & 20.3 & 0.7 \\
\bottomrule
\end{tabular}%
\end{adjustbox}
\end{table}

As shown in Tables~\ref{tab:context-scaling}--\ref{tab:tool-call-scaling}, SAPO-30B scales steadily as the context budget increases and maintains competitive accuracy with lower average token consumption at the 128K limit. Under increasing tool-call budgets, SAPO-30B remains stable, while GRPO-30B and Miro-30B degrade when the interaction budget becomes large. This suggests that SAPO improves not only final accuracy, but also the agent's ability to manage long search trajectories under larger inference budgets.
\endgroup
\FloatBarrier

\section{Data and Artifact Documentation}
\label{app:data-artifacts}

\subsection{Data Screening and Anonymization}
We apply automatic filtering and LLM-as-a-judge verification during data synthesis to remove offensive or unsafe content, and we exclude examples involving private personal identifiers. The released data and artifacts are anonymized and do not include author information or annotator identities.

\subsection{Dataset Scope}
Our data covers both Chinese and English information-seeking tasks. The Chinese data is synthesized from Baidu Baike-style encyclopedic knowledge, covering broad factual domains such as people, organizations, locations, events, and cultural entities; the English data follows a Wikipedia-based synthesis pipeline and covers similar open-domain encyclopedic topics.

\subsection{Licensing and Terms}
We use cited model checkpoints, evaluation benchmarks, and open-source data synthesis methods in accordance with their stated licenses and access terms. For redistributed artifacts, we preserve the required citations and license notices, and release only generated data, code, and prompts that are compatible with research use.

\section{Case Study}
\label{case}

To qualitatively illustrate the differences between SAPO and GRPO, we conducted a comparative case study on the same question using trajectories generated by Qwen3-8B-SAPO and Qwen3-8B-GRPO. As shown in Figure~\ref{fig:goodcase}, SAPO progressively retrieved and cited answer-relevant entities and ultimately answered the question correctly. However, the GRPO trajectory (as shown in Figure~\ref{fig:badcase}) gradually deviated from the answer node and ultimately led to an error.

\begin{figure}[htbp]
    \centering
    \begin{tcolorbox}[
        title={Trajectory Format},
        colframe=darkgray,
        colback=white,
        coltitle=white,
        fonttitle=\large\bfseries,
        arc=1mm,
        boxrule=0.5mm,
        left=8pt, right=8pt, top=8pt, bottom=8pt,
        fontupper=\footnotesize
    ]
    
        \textbf{Question}: A new type of waterproof roofing material is primarily used for waterproofing and decoration. Its development coincides exactly with the founding of a South American football team. This team formerly featured a player whose surname starts with the letter A, and is currently captained by a player whose surname starts with the letter M. What is the name of this material? \\
        \textbf{ER Graph}: [
("Asphalt Shingle", "Type", "New type of waterproof roofing material"),
("Asphalt Shingle", "Usage", "Waterproofing and decoration"),
("Asphalt Shingle", "Development Year", "1893"),
("1893", "Establishment Year", "Argentina National Men's Football Team"),
("Argentina National Men's Football Team", "Associated Player", "Pablo Aimar"),
("Argentina National Men's Football Team", "Current Captain", "Lionel Messi")
] \\
        \textbf{Answer}: Asphalt Shingle
        
        \textcolor{blue}{\texttt{<|im\_start|>assistant}} \\
        \textcolor{blue}{\texttt{<think>}} \\
        First, the question is about a new type of waterproof roofing material... \\
        \textcolor{blue}{\texttt{</think>}} \\
        \textcolor{blue}{\texttt{<tool\_call>}} \\
        \{"name": "search", "arguments": \{"query": ["New waterproof roofing material waterproofing decoration", "Waterproof roofing material development time", "South American football team founding time player surname starts with A captain surname starts with M"]\}\} \\ 
        \textcolor{blue}{\texttt{</tool\_call>}} \textcolor{blue}{\texttt{<|im\_end|>}} \\
        \textcolor{orange}{\texttt{<|im\_start|>user}} \\
        \textcolor{orange}{\texttt{<tool\_response>}} \\
        A Quark search for 'New waterproof roofing material waterproofing decoration' found 5 results:...\textcolor{green!50!black}{Argentina National Men's Football Team}...\textcolor{green!50!black}{Messi}... \\
        \textcolor{orange}{\texttt{</tool\_response>}} \textcolor{orange}{\texttt{<|im\_end|>}} \\
        \textcolor{blue}{\texttt{<|im\_start|>assistant}} \\
        \textcolor{blue}{\texttt{<think>}} \\
        First, from the search results, I found information about the \textcolor{green!50!black}{Argentine national football team}...\textcolor{green!50!black}{Messi}...\textcolor{green!50!black}{1893}... \\
        \textcolor{blue}{\texttt{</think>}} \\
        \textcolor{blue}{\texttt{<tool\_call>}} \\
        \{"name": "search", "arguments": \{"query": ["Asphalt shingles development time 1893", "Fiberglass shingles alias development time", "New waterproof roofing material alias 1893"]\}\} \\ 
        \textcolor{blue}{\texttt{</tool\_call>}} \textcolor{blue}{\texttt{<|im\_end|>}} \\
        \textcolor{orange}{\texttt{<|im\_start|>user}} \\
        \textcolor{orange}{\texttt{<tool\_response>}} \\
        A Quark search for 'Asphalt shingles development time 1893' found 5 results:...\textcolor{green!50!black}{Asphalt Shingles}... \\
        \textcolor{orange}{\texttt{</tool\_response>}} \textcolor{orange}{\texttt{<|im\_end|>}} \\
        \textcolor{blue}{\texttt{<|im\_start|>assistant}} \\
        \textcolor{blue}{\texttt{<think>}} \\
        First, from the search results, I found the key information: fiberglass-reinforced asphalt shingles (also known as \textcolor{green!50!black}{asphalt shingles}) were developed in \textcolor{green!50!black}{1893}...\textcolor{green!50!black}{Messi}... \\
        \textcolor{blue}{\texttt{</think>}} \\
        \textcolor{red}{\texttt{<answer>}} \\
        This type of waterproof shingle material is called \textbf{\textcolor{green!50!black}{asphalt shingle}}. \\
        \textcolor{red}{\texttt{</answer>}} \textcolor{blue}{\texttt{<|im\_end|>}}\textcolor{green!50!black}{\ding{51}}
    
    \end{tcolorbox}
    \caption{A case from the development set where Qwen3-8B-SAPO successfully solves the question.}
    \label{fig:goodcase}
\end{figure}

\begin{tcolorbox}[
        title={Trajectory Format},
        breakable,
        colframe=darkgray,
        colback=white,
        coltitle=white,
        fonttitle=\large\bfseries,
        arc=1mm,
        boxrule=0.5mm,
        left=8pt, right=8pt, top=8pt, bottom=8pt,
        fontupper=\footnotesize
]
    
        \textbf{Question}: A new type of waterproof roofing material is primarily used for waterproofing and decoration. Its development coincides exactly with the founding of a South American football team. This team formerly featured a player whose surname starts with the letter A, and is currently captained by a player whose surname starts with the letter M. What is the name of this material? \\
        \textbf{ER Graph}: [
("Asphalt Shingle", "Type", "New type of waterproof roofing material"),
("Asphalt Shingle", "Usage", "Waterproofing and decoration"),
("Asphalt Shingle", "Development Year", "1893"),
("1893", "Establishment Year", "Argentina National Men's Football Team"),
("Argentina National Men's Football Team", "Associated Player", "Pablo Aimar"),
("Argentina National Men's Football Team", "Current Captain", "Lionel Messi")
] \\
        \textbf{Answer}: Asphalt Shingle
        
        \textcolor{blue}{\texttt{<|im\_start|>assistant}} \\
        \textcolor{blue}{\texttt{<think>}} \\
        First, the question concerns a new type of waterproof shingle material, primarily used for waterproofing and decoration... \\
        \textcolor{blue}{\texttt{</think>}} \\
        \textcolor{blue}{\texttt{<tool\_call>}} \\
        \{"name": "search", "arguments": \{"query": ["New type of waterproof roofing tile", "South American football team player's surname starts with A", "South American football team player's surname starts with M"]\}\} \\ 
        \textcolor{blue}{\texttt{</tool\_call>}} \textcolor{blue}{\texttt{<|im\_end|>}} \\
        \textcolor{orange}{\texttt{<|im\_start|>user}} \\
        \textcolor{orange}{\texttt{<tool\_response>}} \\
        A Quark search for 'New type of waterproof roofing tile' found 5 results:...\textcolor{green!50!black}{Argentina National Men's Football Team}...\textcolor{green!50!black}{Messi}...\textcolor{red!50!black}{Brazil National Team}...\textcolor{red!50!black}{Marquinhos}... \\
        \textcolor{orange}{\texttt{</tool\_response>}} \textcolor{orange}{\texttt{<|im\_end|>}} \\
        \textcolor{blue}{\texttt{<|im\_start|>assistant}} \\
        \textcolor{blue}{\texttt{<think>}} \\
        I have reviewed the search results. \textcolor{red!50!black}{The Brazil team} was founded in 1914. Its captain is \textcolor{red!50!black}{Marquinhos} (starts with M) and... \\
        \textcolor{blue}{\texttt{</think>}} \\
        \textcolor{blue}{\texttt{<tool\_call>}} \\
        \{"name": "search", "arguments": \{"query": ["Waterproof roofing material developed in 1914", "roofing material invented 1914"]\}\} \\ 
        \textcolor{blue}{\texttt{</tool\_call>}} \textcolor{blue}{\texttt{<|im\_end|>}} \\
        \textcolor{orange}{\texttt{<|im\_start|>user}} \\
        \textcolor{orange}{\texttt{<tool\_response>}} \\
        A Quark search for 'Waterproof roofing material developed in 1914' found 5 results:...\textcolor{red!50!black}{1914}...\textcolor{red!50!black}{modified bitumen technologies}... \\
        \textcolor{orange}{\texttt{</tool\_response>}} \textcolor{orange}{\texttt{<|im\_end|>}} \\
        \textcolor{blue}{\texttt{<|im\_start|>assistant}} \\
        \textcolor{blue}{\texttt{<think>}} \\
        The year is 1914, matching the founding of the Brazil team. The material developed around this time is \textcolor{red!50!black}{Modified Bitumen Membrane} (or Bituminous Membrane). \\
        \textcolor{blue}{\texttt{</think>}} \\
        \textcolor{red}{\texttt{<answer>}} \\
        The name of this material is \textcolor{red!50!black}{Modified Bitumen Membrane}. \\
        \textcolor{red}{\texttt{</answer>}} \textcolor{blue}{\texttt{<|im\_end|>}}\textcolor{red!50!black}{\ding{55}}
    
\end{tcolorbox}
\captionof{figure}{A case from the development set where Qwen3-8B-GRPO fails to solve the question due to insufficient citation of retrieved entities.}
\label{fig:badcase}

\end{document}